\documentclass[11pt]{article}

\usepackage[preprint]{acl}
\usepackage{times}
\usepackage{latexsym}

\usepackage[T1]{fontenc}
\usepackage[utf8]{inputenc}

\usepackage{microtype}

\usepackage{inconsolata}

\usepackage{graphicx}
\usepackage{inconsolata}
\usepackage{arydshln}
\usepackage{hyperref}
\usepackage{amsmath}
\usepackage{algorithmic}
\usepackage{textcomp}
\usepackage{arydshln}
\usepackage{geometry} %%%
\usepackage{multirow}
\usepackage{booktabs}
\usepackage{amsfonts}
\usepackage{graphicx}
\usepackage{pifont}
\usepackage{color}
\usepackage[dvipsnames]{xcolor}
\usepackage{colortbl}
\usepackage{xspace}
\usepackage{lipsum}
\usepackage{caption}
\usepackage{subcaption}
\usepackage{enumitem}
\usepackage{bm}
\usepackage{adjustbox}
\usepackage{tabularray}
\usepackage{array}

\usepackage{listings}
\usepackage{xcolor}

\lstdefinestyle{promptstyle}{
    basicstyle=\ttfamily\footnotesize,
    breaklines=true,                  
    frame=single,                     
    rulecolor=\color{black!30},
    backgroundcolor=\color{gray!5},
    columns=fullflexible,
    keepspaces=true,
    breakindent=0pt,
    extendedchars=true,
    literate={-}{-}1,
}

\usepackage{graphbox}

\newcommand{\greencheck}{\textcolor{OliveGreen}{\ding{52}}}
\newcommand{\redcross}{\textcolor{BrickRed}{\ding{56}}}
\newcommand{\model}[0]{{MMErroR}\xspace}
\newcommand{\num}[0]{{1,997}\xspace}
\definecolor{lightgreen}{RGB}{212, 233, 208}

\usepackage[table]{xcolor}

\definecolor{TopOne}{RGB}{206,220,204}
\definecolor{TopTwo}{RGB}{223,233,221}
\definecolor{TopThree}{RGB}{239,245,238}

\title{MMErroR: A Benchmark for Erroneous Reasoning \\in Vision-Language Models}
\author{
    \href{https://orcid.org/0009-0009-3928-7495}{\textcolor{black}{Yang Shi}}$^{1}$\thanks{These authors contributed equally to this work.},
    Yifeng Xie$^{2 *}$,
    Minzhe Guo$^{1}$,
    \href{https://orcid.org/0009-0006-2839-3901}{\textcolor{black}{Liangsi Lu}}$^{1}$,
    Mingxuan Huang$^{3}$, \\
    \textbf{
    Jingchao Wang$^{4}$,
    Zhihong Zhu$^{4}$,
    Boyan Xu$^{1}$\thanks{Corresponding author.},
    Zhiqi Huang$^{4}$} \\
    $^{1}$Guangdong University of Technology \quad
    $^{2}$Hong Kong Baptist University \\
    $^{3}$Sun Yat-sen University  \quad
    $^{4}$Peking University \\
    \texttt{\{sudo.shiyang, evfxie, capynt, lu.liangsi.cn\}@gmail.com} \\
    \texttt{huangmx53@mail2.sysu.edu.cn} \quad
    \texttt{ethanwangjc@163.com} \\
    \texttt{zhihongzhu@stu.pku.edu.cn} \quad
    \texttt{hpakyim@gmail.com} \quad
    \texttt{zhiqihuang@pku.edu.cn} \\
}

\begin{document}
\maketitle
\begin{abstract}
Recent advances in Vision-Language Models (VLMs) have improved performance in multi-modal learning, raising the question of whether these models truly understand the content they process. Crucially, can VLMs detect when a reasoning process is wrong and identify its error type?
To answer this, we present MMErroR, a multi-modal benchmark of 1997 samples, each embedding a single coherent reasoning error. These samples span 24 subdomains across six top-level domains, ensuring broad coverage and taxonomic richness. Unlike existing benchmarks that focus on answer correctness, MMErroR targets a process-level, error-centric evaluation that requires models to detect incorrect reasoning and classify the error type within both visual and linguistic contexts.
We evaluate 12 representative VLMs, and even the best model, Gemini-3-Pro-Preview, classifies the error correctly in only 66.65\% of cases, underscoring the challenge of identifying erroneous reasoning. Furthermore, the ability to accurately identify errors offers valuable insights into the capabilities of multi-modal models.
Project Page: \href{https://mmerror-benchmark.github.io}{https://mmerror-benchmark.github.io}
% Our code and dataset are available at https://github.com/xx.
\end{abstract}
\begin{figure}[t]
\centering
\renewcommand{\arraystretch}{1.2}
\resizebox{1\linewidth}{!}{
% \Large
\begin{tabular}{l ccc}
\toprule
\textbf{Benchmarks} & \textbf{Multi-Modality} & \textbf{Multi-Domain} & \textbf{Categorize} \\
\midrule
ProcessBench~\cite{zheng2025processbench} & \redcross & \redcross & \redcross \\
PRISM-Bench~\cite{fang2025flux}  &  \greencheck  & \redcross & \redcross  \\
ErrorRadar~\cite{yan2024errorradar}   &  \greencheck  & \redcross & \redcross \\
\cdashline{1-4}
\model (ours)           & \greencheck  & \greencheck & \greencheck \\
\bottomrule
\\
\end{tabular}
}
\renewcommand{\arraystretch}{1}
\includegraphics[width=0.9\linewidth]{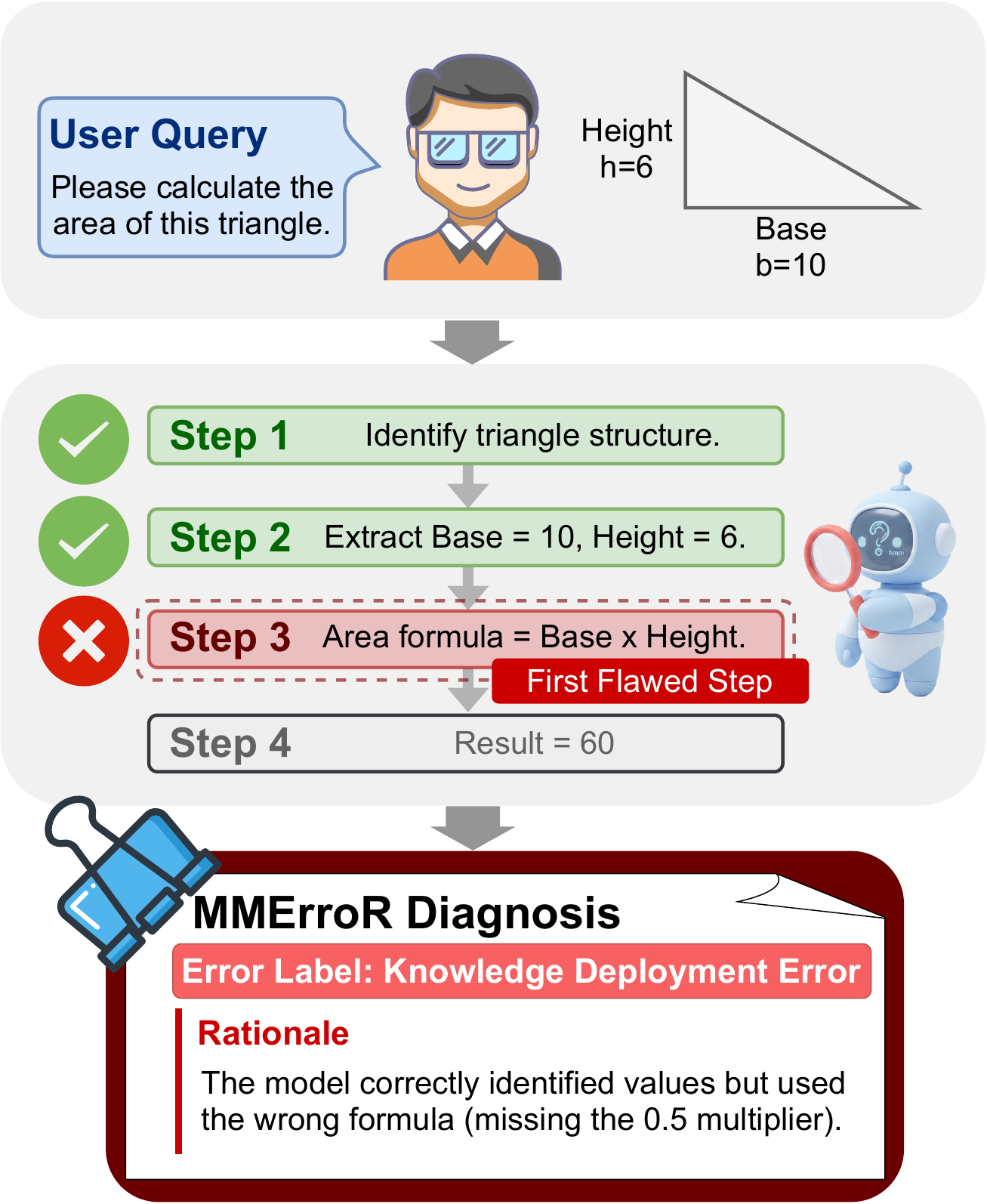}
\caption{Comparison with existing error localization benchmarks. A sample from \model illustrates an erroneous reasoning chain where the model is required to both detect and classify the error type.}
% \caption{\textbf{(Top)} Comparison with existing error localization benchmarks. \textbf{(Bottom)} The sample from \model illustrates an error reasoning for which the model is required to both detect and classify the error type.}
\label{fig:intro}
\end{figure}

\section{Introduction}
The rapid advancement of large multi-modal models has led to substantial progress in unified reasoning across vision and language, pushing performance~\cite{alayrac2022flamingo,team2023gemini} on various multi-modal tasks closer to or surpassing in certain benchmarks~\cite{hurst2024gpt,yue2024mmmu}. These improvements create an impression that large multi-modal models are approaching a robust, human-like understanding of cross-modal content, a perception further reinforced by their growing deployment in real-world applications such as educational assistants, medical imaging analysis, and autonomous systems~\cite{liu2023visual,tu2024towards,zitkovich2023rt}.

\begin{figure*}[t]
\centering
\includegraphics[width=0.9\linewidth]{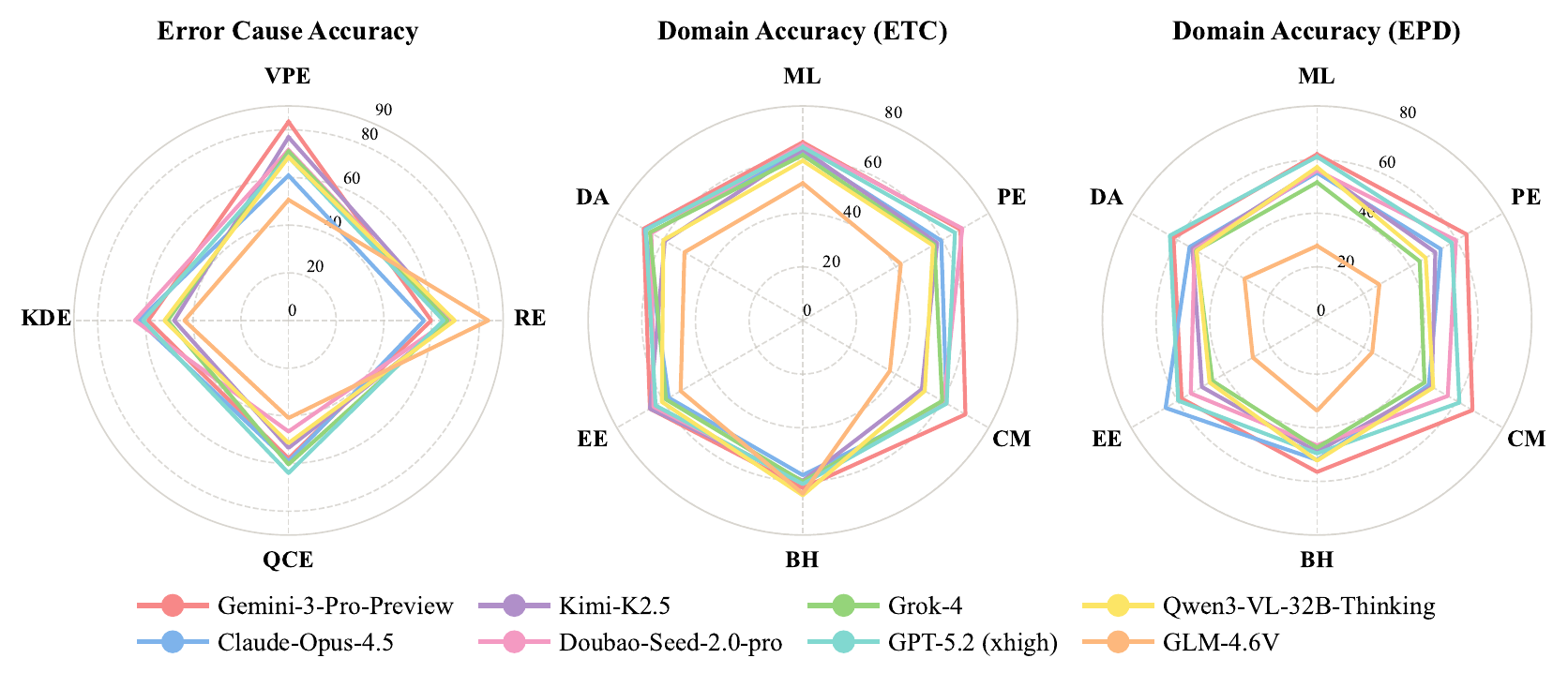}
\caption{Comparison of different VLMs across various task domains and four error types: Visual Perception Error (VPE), Reasoning Error (RE), Question Comprehension Error (QCE), and Knowledge Deployment Error (KDE).}
\label{fig:score}
\end{figure*}

Despite this progress, a fundamental question remains: Do these models genuinely understand the meaning between visual and textual content, or are they merely generating statistically plausible yet superficial associations through pattern matching? Moreover, if presented with an erroneous reasoning chain about the same multi-modal scene, can the model not only detect the error but also pinpoint its cause and type?
As shown in Figure~\ref{fig:intro}, existing benchmarks for error localization focus primarily on identifying which step in the reasoning process is incorrect, offering limited insight into the nature of the failure. In contrast, classifying the error type enables a diagnostic understanding of why the model went astray: whether due to a breakdown in visual grounding, a logical inconsistency, a factual hallucination, or a computational mistake. Each type of error reflects a distinct weakness in the model’s multimodal comprehension pipeline. Thus, a deep evaluation of a model's ability to diagnose reasoning errors serves as a litmus test for genuine multi-modal understanding.

To address this gap, we introduce \model (\textbf{M}ulti-\textbf{M}odal \textbf{Erro}r \textbf{R}easoning Benchmark), a benchmark designed to evaluate the VLM's ability to diagnose multi-modal erroneous reasoning. \model comprises \num meticulously curated samples distributed across several core reasoning domains: Data \& Analytics (DA), Physics \& Engineering (PE), Chemistry \& Materials (CM), Earth \& Environment (EE), Biology \& Healthcare (BH), and Mathematics \& Logic (ML). Every sample contains a coherent Chain-of-Thought~\cite{wei2022chain} into which a representative error has been injected. To enable unambiguous root-cause attribution, each chain is constructed to contain exactly one error. The models are required to determine whether to invoke an error label and, if so, identify its precise type. This controlled design yields fine-grained insights into model weaknesses and should be interpreted as a stress-test of error diagnosis rather than a full calibration benchmark over a natural mixture of correct and incorrect chains.

To ensure a rigorous assessment, we design two distinct evaluation modes: Error Type Classification (ETC) and Error Presence Detection (EPD). In the first mode, we explicitly inform the model that an error exists and prompt it to classify the error type. In the second mode, the model is required to first decide whether to invoke an error label before diagnosing it. Because the current release contains only erroneous reasoning chains, this second setting should be interpreted as a controlled stress-test of error sensitivity and attribution rather than a full calibration benchmark.
As shown in Figure~\ref{fig:score}, the evaluation of 12 representative VLMs reveals that these tasks remain challenging. Even the most capable model in our study, Gemini-3-Pro-Preview, successfully identifies the error type in only 66.65\% of cases, with performance on fine-grained error classification remaining substantially below human performance.
This result underscores a notable gap between the generative capability of current models and their capacity for introspective verification.

In summary, our key contributions are as follows:
\begin{itemize}
\item We propose \model, a benchmark designed specifically for error-type evaluation of multi-modal reasoning, enabling fine-grained assessment of whether models can detect and diagnose flawed reasoning in vision-language contexts.
\item Through a comprehensive empirical evaluation of 12 representative VLMs, we reveal that current models struggle significantly with introspective error detection and classification, uncovering a critical gap in their ability to achieve trustworthy self-oversight in multi-modal reasoning.
\item We conduct in-depth diagnostic analysis to uncover key factors influencing erroneous reasoning in multi-modal learning, such as modality misalignment, logical inconsistency, and perceptual over-reliance, providing actionable insights for future model improvement.
\end{itemize}

\section{\model}

\subsection{Task Classification}
In \model, we design two complementary evaluation tasks to assess a model’s ability
to detect and diagnose errors in multi-modal reasoning processes.
Together, these tasks evaluate whether a model can explicitly invoke an error label under controlled erroneous conditions and, if so, correctly identify its underlying cause.

\paragraph{Error Type Classification (ETC)}
Given an image, a corresponding question, and a complete reasoning chain that is guaranteed
to contain exactly one error, the model is required to identify the specific error type
from a predefined taxonomy.
The error types include:
\emph{Visual Perception Error}, involving incorrect visual grounding such as object
misidentification, misinterpretation of spatial relations, or erroneous reading of symbols
and diagrams;
\emph{Knowledge Deployment Error}, arising from misuse or misapplication of external
knowledge, such as incorrect physical laws, mathematical formulas, or domain-specific concepts;
\emph{Question Comprehension Error}, caused by misunderstanding the intent of the question,
overlooking key constraints, or incorrectly interpreting the required target;
and \emph{Reasoning Error}, which includes logical fallacies, missing premises,
invalid inference steps, or internal inconsistencies in the reasoning process.

\paragraph{Error Presence Detection (EPD)}
Under the same input setting, the model must first determine whether to select ``No Error'' or ``Error Present'' for the provided reasoning chain.
If the model predicts that an error is present, it must then determine the type of the error.
Because the current release of \model contains only erroneous reasoning chains, EPD is intended as a controlled stress-test of error sensitivity and attribution rather than a full calibration benchmark on mixed correct and incorrect chains.

\subsection{Benchmark Construction}
In this subsection, we detail the construction of \model. The process is organized into four main steps.

\begin{figure*}[tb]
  \centering
  \includegraphics[width=0.55\linewidth, align=c]{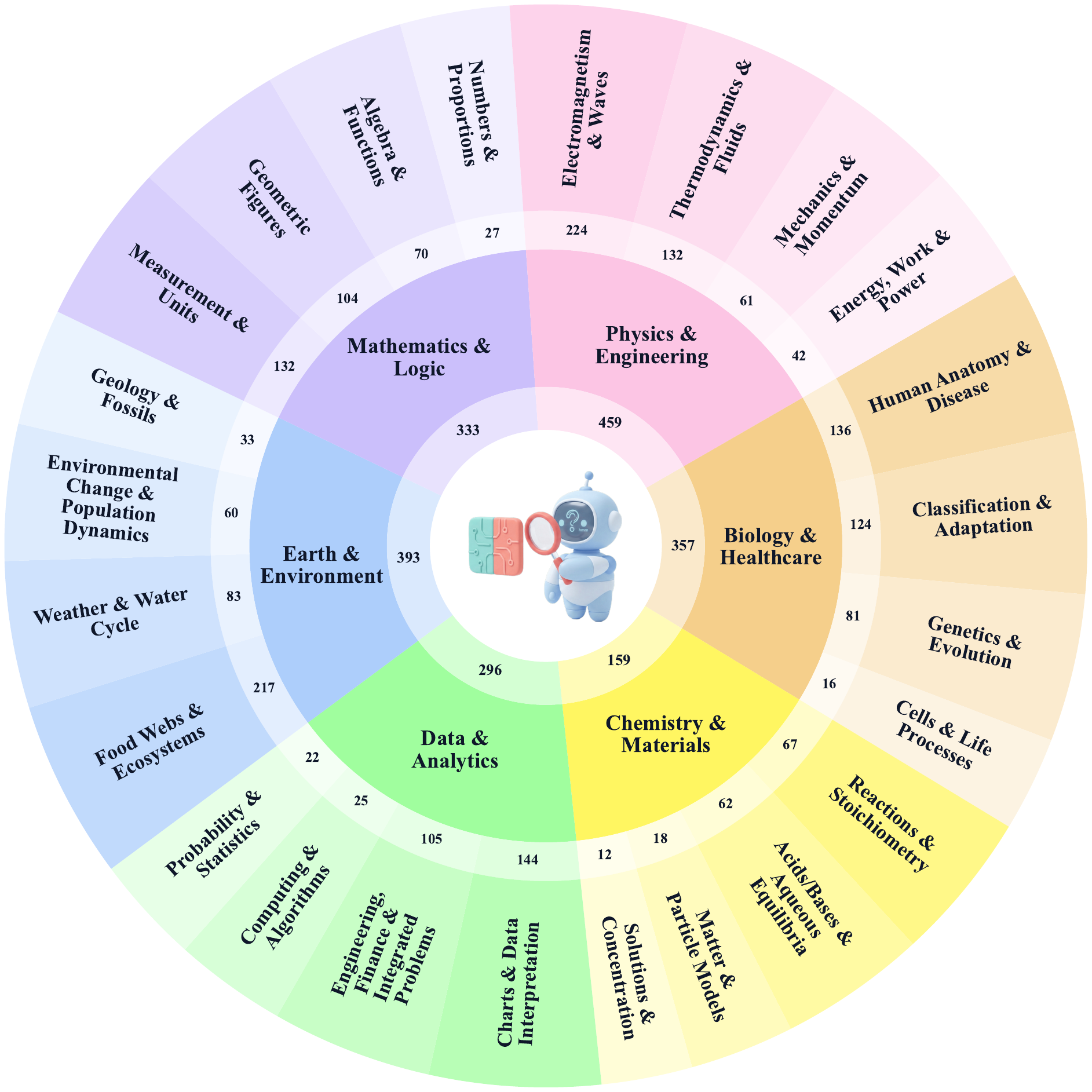}
  \includegraphics[width=0.43\linewidth, align=c]{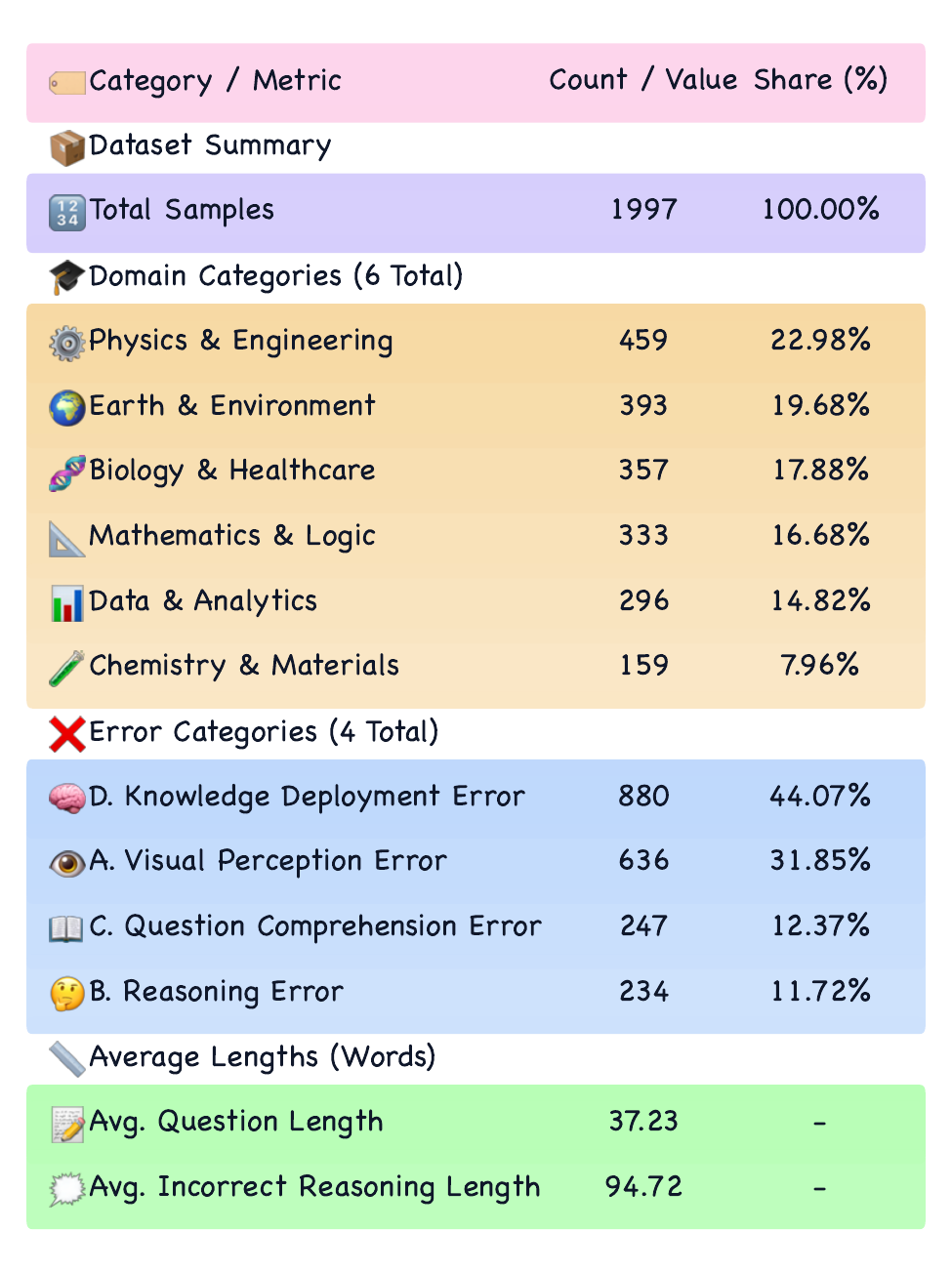}
  \caption{Detailed analysis of the domain, subdomain, and error-type statistics of \model.}
  \label{fig:detail}
\end{figure*}

\paragraph{Problem Curation}
To ensure both broad domain coverage and targeted evaluation of multi-modal reasoning,
\model sources its initial image--question--answer triplets from a set of established benchmarks,
including MMMU~\cite{yue2024mmmu}, MathVista~\cite{lu2023mathvista}, MathVerse~\cite{zhang2024mathverse},
ScienceQA~\cite{lu2022learn}, and AI2D~\cite{kembhavi2016diagram}.
These benchmarks are widely adopted in vision--language evaluation and remain challenging
for current models, providing a reliable foundation for constructing non-trivial reasoning instances.

To avoid over-representation of any single domain, we apply stratified sampling to balance
the number of instances across domains.
In addition, we perform a complexity-aware filtering step that removes overly simple
or low-information samples, retaining only instances that require multi-step reasoning
and substantive cross-modal inference.
This design ensures that \model emphasizes challenging reasoning scenarios rather than
surface-level perception or pattern matching.
Details of the filtering procedure are provided in Appendix~\ref{sec:appendix_1}.

% 看数据合成的论文怎么讲novelty的
\paragraph{Error Injection}
To construct erroneous reasoning chains while maintaining control and realism,
we adopt a hybrid generation strategy.
For each curated instance, GPT-5~\cite{openai:gpt-5-system-card} is used to inject
a single, contextually coherent error into an otherwise plausible reasoning chain,
under explicit generation constraints (see Appendix~\ref{sec:appendix_2}).
We intentionally enforce a single root-cause error per chain to obtain unambiguous diagnostic labels.
Although real-world reasoning failures may involve cascaded mistakes, allowing multiple interacting errors would substantially confound attribution.
The injected errors are restricted to one of four predefined categories:
Visual Perception Error, Knowledge Deployment Error,
Question Comprehension Error, and Reasoning Error.
Aside from the injected error, the remaining reasoning steps are required
to be locally coherent and logically valid, ensuring that each instance reflects
a realistic and non-trivial reasoning failure.

\paragraph{Data Verification}
To ensure the quality and realism of the generated erroneous reasoning chains,
we employ a rigorous three-round human verification protocol.
We invited a total of 20 experts (including 6 professors in the corresponding domains and 14 doctoral students) to conduct a 23-day inspection on the initial 10,000 samples. During this period, we ensured that each sample was inspected by three different experts in three separate rounds.
A reasoning chain is discarded if it satisfies any of the following conditions:
(1) the erroneous reasoning is incoherent or irrelevant to the original question;
(2) the assigned error type is incorrect;
(3) the error is ambiguous or plausibly attributable to multiple error categories.
Only samples with unanimous approval are retained, resulting in 3,929 valid instances in Round 1, 3,239 in Round 2, and 3,148 instances after the third round.
The marginal elimination rate of 2.81\% in the final round reflects an observed agreement of 97.19\%~\cite{artstein2008inter}, suggesting annotation stability.

\paragraph{Quality Assurance}
To further ensure the quality and realism of erroneous reasoning chains in \model,
we apply an additional human scoring and filtering stage.
Each generated reasoning chain is independently evaluated by at least two linguistics experts along four quality dimensions:
\emph{Coherence}, \emph{Step Clarity}, \emph{Error Localizability}, and
\emph{Semantic Consistency}.
Each dimension is rated on a three-point scale:
$-1$ (unsatisfactory), $0$ (adequate), and $1$ (satisfactory).
A reasoning chain is retained only if its average score across evaluators exceeds a predefined threshold of $0.5$.
This criterion ensures that retained samples exhibit both a realistic reasoning flow and a well-localized, non-trivial error.
After this scoring-based filtering, a total of \num high-quality erroneous reasoning samples are retained for final inclusion.
This quality assurance pipeline ensures that \model is both challenging and reliable for benchmarking multi-modal error detection and diagnosis.
Furthermore, a rigorous pilot study on a stratified sample of 300 instances achieved a Cohen’s Kappa of $\kappa=0.796$~\cite{cohen1960coefficient}. These metrics verify the high consistency of our annotation standards.

\subsection{Data Analysis}
Figure~\ref{fig:detail} summarizes the hierarchical composition of \model. Among the six top-level domains, Physics \& Engineering accounts for the largest portion of the dataset (22.98\%, 459 samples), followed by Earth \& Environment (19.68\%, 393 samples), Biology \& Healthcare (17.88\%, 357 samples), Mathematics \& Logic (16.68\%, 333 samples), and Data \& Analytics (14.82\%, 296 samples), while Chemistry \& Materials constitutes 7.96\% (159 samples). At the subdomain level, the benchmark remains distributed across 24 categories, with Electromagnetism \& Waves, Food Webs \& Ecosystems, Charts \& Data Interpretation, and Human Anatomy \& Disease among the largest groups. At the error-type level, Knowledge Deployment Error is the most prevalent (44.07\%, 880 samples), followed by Visual Perception Error (31.85\%, 636 samples), Question Comprehension Error (12.37\%, 247 samples), and Reasoning Error (11.72\%, 234 samples). On average, questions contain 37.23 words, and erroneous reasoning chains average 94.72 words, indicating non-trivial reasoning contexts with multiple intermediate steps. Overall, this balanced yet challenge-oriented distribution enables \model to cover diverse multi-modal scenarios while focusing on process-level reasoning failures rather than superficial answer mistakes.

\begin{table*}[!t]
\centering
\resizebox{1\linewidth}{!}{
\begin{tabular}{l c c c c c c c c}
\toprule
 & \textbf{ML} & \textbf{PE} & \textbf{CM} & \textbf{BH} & \textbf{EE} & \textbf{DA} & \textbf{Macro} & \textbf{Overall} \\
\midrule
\rowcolor{gray!10}
\multicolumn{9}{c}{\textbf{Baselines}} \\
Random Choice & 22.10 & 23.62 & 24.18 & 24.06 & 21.50 & 25.53 & 23.50 & 23.45 \\
Human Expert (Low)  & 78.33 & 75.63 & 73.75 & 77.09 & 74.70 & 76.85 & 76.06 & 76.22 \\
Human Expert (High) & 91.07 & 88.65 & 87.50 & 90.15 & 88.96 & 90.18 & 89.42 & 89.52 \\
\midrule
\rowcolor{gray!10}
\multicolumn{9}{c}{\textbf{Open-weights Vision-Language Models}} \\
LLaMA-4-Scout & 43.84 & 34.86 & 39.62 & 49.02 & 28.24 & 42.23 & 39.64 & 39.06 \\
LLaMA-4-Maverick & 42.64 & 36.17 & 39.62 & 50.14 & 28.50 & 42.57 & 39.94 & 39.46 \\
Qwen3-VL-32B-Instruct & 47.45 & 32.68 & 28.93 & 58.54 & 36.64 & 51.35 & 42.60 & 43.01 \\
GLM-4.6V & 51.05 & 41.18 & 37.11 & 63.87 & 52.16 & 51.35 & 49.45 & 50.23 \\
Qwen3-VL-32B-Thinking & 59.46 & 54.90 & 52.20 & \cellcolor{TopTwo}65.83 & 60.81 & 59.80 & 58.83 & 59.29 \\
Kimi-K2.5 & 63.66 & 55.56 & 51.57 & 58.82 & \cellcolor{TopOne}66.67 & 61.15 & 59.57 & 60.19 \\
\midrule
\rowcolor{gray!10}
\multicolumn{9}{c}{\textbf{Proprietary Vision-Language Models}} \\
Qwen-VL-Max & 53.45 & \cellcolor{TopThree}65.36 & \cellcolor{TopTwo}69.18 & \cellcolor{TopOne}66.39 & 42.75 & 57.09 & 59.04 & 58.19 \\
Grok-4 & 61.56 & 55.56 & 59.75 & 61.34 & 59.29 & 65.88 & 60.56 & 60.19 \\
Claude-Opus-4.5 & 62.76 & 61.00 & 61.64 & 57.70 & 56.74 & \cellcolor{TopThree}68.58 & 61.40 & 61.04 \\
GPT-5.2 (xhigh) & \cellcolor{TopThree}64.56 & 63.62 & \cellcolor{TopThree}62.26 & 60.50 & 65.14 & \cellcolor{TopOne}69.59 & \cellcolor{TopThree}64.28 & \cellcolor{TopThree}64.30 \\
Doubao-Seed-2.0-pro & \cellcolor{TopTwo}65.47 & \cellcolor{TopOne}67.32 & 61.01 & 59.94 & \cellcolor{TopTwo}66.16 & 66.22 & \cellcolor{TopTwo}64.35 & \cellcolor{TopTwo}64.80 \\
Gemini-3-Pro-Preview & \cellcolor{TopOne}66.37 & \cellcolor{TopTwo}66.88 & \cellcolor{TopOne}69.81 & \cellcolor{TopThree}64.43 & \cellcolor{TopThree}65.39 & \cellcolor{TopTwo}69.26 & \cellcolor{TopOne}67.02 & \cellcolor{TopOne}66.65 \\
\bottomrule
\end{tabular}
}
\caption{Accuracy (\%) comparison of baselines under ETC evaluation. The best, second-best, and third-best vision-language models in each column are highlighted by \cellcolor{TopOne}{dark}, \cellcolor{TopTwo}{medium}, and \cellcolor{TopThree}{light} background shades, respectively. Baseline methods and human experts are excluded from ranking. Within the open-weights and proprietary groups, models are ordered by Overall accuracy in ascending order.}
\label{tab:main_results}
\end{table*}

\begin{figure*}[t]
    \centering
    \includegraphics[width=0.98\linewidth]{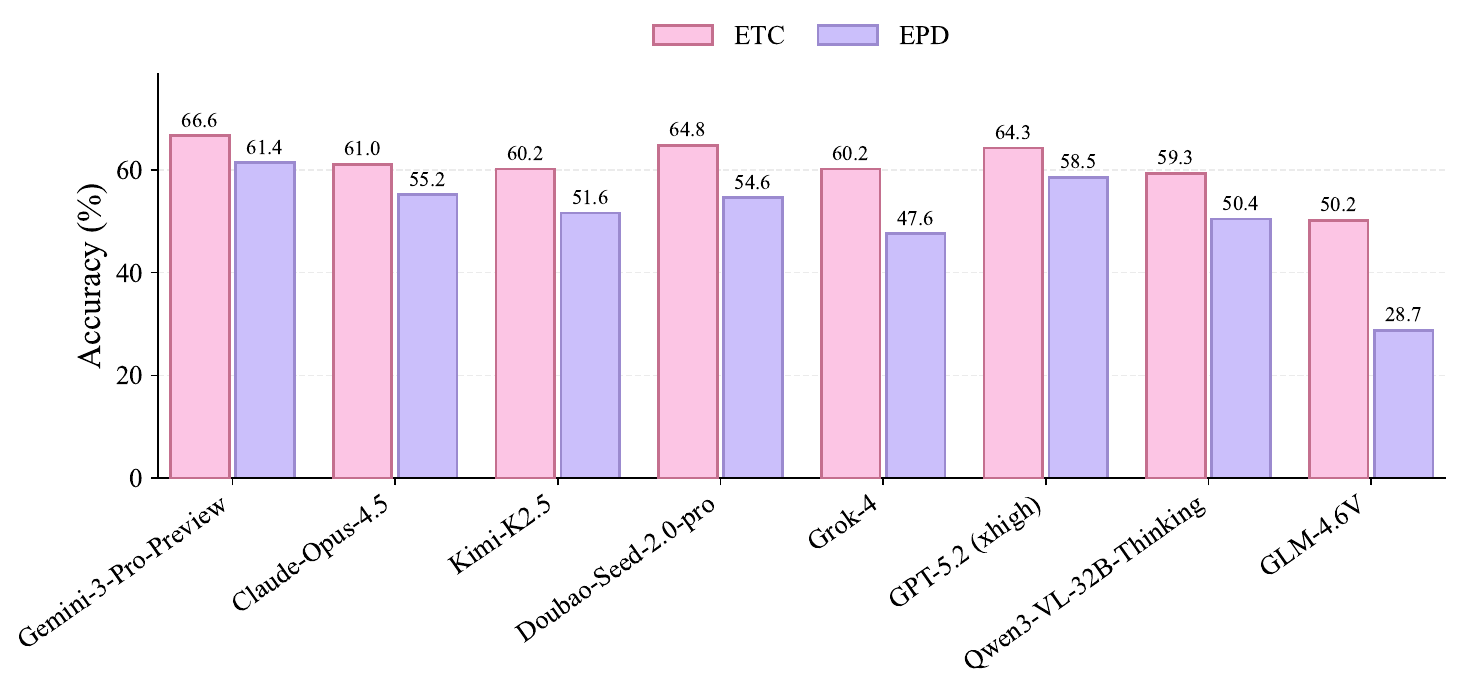}
    \caption{Performance comparison of different VLMs on \model. We evaluate and compare performance under two settings: Error Type Classification (ETC) and Error Presence Detection (EPD).}
\label{fig:model_scores}
\end{figure*}

\section{Experiment Settings}
\subsection{Models}
We evaluate a representative set of 12 VLMs and organize them into two groups according to model accessibility, namely open-weights models and proprietary models. The open-weights group includes LLaMA-4-Scout~\cite{meta:llama-4-scout-card}, LLaMA-4-Maverick~\cite{meta:llama-4-maverick-card}, Qwen3-VL-32B-Instruct~\cite{qwen:qwen3-vl-report}, Qwen3-VL-32B-Thinking~\cite{qwen:qwen3-vl-report}, GLM-4.6V~\cite{hong2025glm}, and Kimi-K2.5~\cite{team2026kimi}. The proprietary group includes Qwen-VL-Max~\cite{alibaba:qwen-vl-max-doc}, Grok-4~\cite{xai:grok-4}, Claude-Opus-4.5~\cite{anthropic:claude-opus-4.5-system-card}, GPT-5.2 (xhigh)~\cite{openai:gpt-5.2-doc}, Doubao-Seed-2.0-pro~\cite{bytedance:seed-2.0-model-card}, and Gemini-3-Pro-Preview~\cite{google:gemini-3-doc}. In addition to these models, we report results for simple baselines (Random Choice) and for Human Expert (Low/High) performance to assess the difficulty of \model.

\subsection{Implementation and Metrics}
We assess model performance using two complementary evaluation protocols: Error-Type Classification (ETC) and Error Presence Detection (EPD).
In both settings, each evaluation instance consists of an image, a question, and a step-by-step reasoning chain.
For the ETC task, the chain is guaranteed to contain exactly one error, and the model must identify its type from four predefined categories.
For the EPD task, the model must first choose between ``No Error'' and ``Error Present'' and, if it predicts an error, further classify its type.
We emphasize that, because \model contains only erroneous reasoning chains, EPD does not measure false-positive behavior on verified clean chains and is not intended as a full calibration benchmark.
Instead, it serves as a controlled stress-test of error sensitivity and attribution under uniformly erroneous conditions.
An always-error strategy does not trivially solve EPD, because credit is awarded only when the model also identifies the correct error type; such a strategy therefore reduces to ETC-level performance.

We adopt a multiple-choice format. Models are prompted to output the label corresponding to their judgment. To provide a fine-grained analysis, we report performance across six distinct dimensions:  Data \& Analytics (DA), Physics \& Engineering (PE), Chemistry \& Materials (CM), Earth \& Environment (EE), Biology \& Healthcare (BH), and Mathematics \& Logic (ML). We also report the Macro Average Score (Macro) across these categories and the Overall Weighted Accuracy (Overall). 
To ensure deterministic and reproducible comparisons, we set the decoding temperature to 0 for all evaluations.

\section{Empirical Results and Analysis}

\subsection{ETC Evaluation Results}
We evaluate performance using the Error Type Classification (ETC) task. As shown in Table~\ref{tab:main_results}, the following observations can be made:

(1) Proprietary models achieve the strongest overall performance on ETC. Gemini-3-Pro-Preview attains the best overall accuracy at 66.65\%, followed by Doubao-Seed-2.0-pro at 64.80\% and GPT-5.2 (xhigh) at 64.30\%. Despite this progress, the gap to Human Expert performance remains substantial, indicating that fine-grained diagnosis of erroneous reasoning is still challenging for current VLMs.

(2) Performance is heterogeneous across domains rather than dominated by a single model. Gemini-3-Pro-Preview achieves the best results in ML at 66.37\% and in CM at 69.81\%, Doubao-Seed-2.0-pro leads PE at 67.32\%, Qwen-VL-Max performs best in BH at 66.39\%, Kimi-K2.5 achieves the top score in EE at 66.67\%, and GPT-5.2 (xhigh) leads DA at 69.59\%. This pattern suggests that error diagnosis in \model depends on multiple underlying capabilities, including domain knowledge, visual grounding, and procedural reasoning.

(3) Among open-weights models, performance also varies substantially. Kimi-K2.5 attains the strongest overall result at 60.19\%, followed by Qwen3-VL-32B-Thinking at 59.29\%, and both remain competitive with several proprietary models in selected domains. In contrast, LLaMA-4-Scout, LLaMA-4-Maverick, and Qwen3-VL-32B-Instruct remain clearly below the leading group, showing that \model provides meaningful discrimination across model families and capability levels.

\begin{table*}[t]
\centering
\resizebox{1\linewidth}{!}{
\begin{tabular}{l c c c c c c c c}
\toprule
 & \textbf{ML} & \textbf{PE} & \textbf{CM} & \textbf{BH} & \textbf{EE} & \textbf{DA} & \textbf{Macro} & \textbf{Overall} \\
\midrule
\rowcolor{gray!10}
\multicolumn{9}{c}{\textbf{Open-weights Vision-Language Models}} \\
LLaMA-4-Maverick & 23.12 & 16.12 & 7.55 & 17.09 & 12.21 & 30.41 & 17.75 & 18.13 \\
LLaMA-4-Scout & 24.62 & 18.74 & 15.09 & 18.21 & 13.23 & 28.72 & 19.77 & 19.73 \\
GLM-4.6V & 27.93 & 27.23 & 23.27 & 35.01 & 25.45 & 31.76 & 28.44 & 28.74 \\
Qwen3-VL-32B-Instruct & 40.84 & 36.17 & 27.04 & 44.54 & 27.99 & 41.55 & 36.36 & 36.91 \\
Qwen3-VL-32B-Thinking & \cellcolor{TopThree}57.06 & 46.84 & 49.69 & \cellcolor{TopThree}54.06 & 44.27 & 52.70 & 50.77 & 50.43 \\
Kimi-K2.5 & 56.46 & 48.58 & 49.69 & 51.54 & 49.11 & \cellcolor{TopThree}55.41 & 51.80 & 51.63 \\
\midrule
\rowcolor{gray!10}
\multicolumn{9}{c}{\textbf{Proprietary Vision-Language Models}} \\
Qwen-VL-Max & 39.04 & 42.05 & 42.14 & 47.34 & 30.28 & 40.54 & 40.23 & 39.96 \\
Grok-4 & 51.65 & 43.57 & 45.91 & 50.14 & 42.75 & 53.38 & 47.90 & 47.57 \\
Doubao-Seed-2.0-pro & 56.16 & \cellcolor{TopTwo}58.17 & \cellcolor{TopThree}56.60 & 49.86 & 53.44 & 53.38 & \cellcolor{TopThree}54.60 & 54.58 \\
Claude-Opus-4.5 & 55.26 & 50.76 & 47.80 & \cellcolor{TopTwo}55.74 & \cellcolor{TopOne}62.85 & 55.07 & 54.58 & \cellcolor{TopThree}55.18 \\
GPT-5.2 (xhigh) & \cellcolor{TopTwo}60.96 & \cellcolor{TopThree}55.99 & \cellcolor{TopTwo}61.64 & 51.26 & \cellcolor{TopTwo}59.29 & \cellcolor{TopOne}65.88 & \cellcolor{TopTwo}59.17 & \cellcolor{TopTwo}58.54 \\
Gemini-3-Pro-Preview & \cellcolor{TopOne}61.86 & \cellcolor{TopOne}63.83 & \cellcolor{TopOne}66.67 & \cellcolor{TopOne}59.94 & \cellcolor{TopThree}56.49 & \cellcolor{TopTwo}62.50 & \cellcolor{TopOne}61.88 & \cellcolor{TopOne}61.39 \\
\bottomrule
\end{tabular}
}
\caption{Accuracy (\%) comparison under EPD evaluation. The best, second-best, and third-best vision-language models in each column are highlighted by dark, medium, and light background shades, respectively. Within the open-weights and proprietary groups, models are ordered by Overall accuracy in ascending order.}
\label{tab:5label}
\end{table*}
\subsection{EPD Evaluation Results}
The Error Presence Detection (EPD) task presents a more challenging setting than the ETC task, requiring models to first determine whether to invoke an error label before attempting to classify the error type. As shown in Figure~\ref{fig:model_scores} and Table~\ref{tab:5label}, all models exhibit a clear performance drop under EPD relative to ETC, while the overall ranking remains broadly consistent. Gemini-3-Pro-Preview attains the best overall accuracy at 61.39\% and the best macro-average at 61.88\%, followed by GPT-5.2 (xhigh) with 58.54\% overall accuracy and Claude-Opus-4.5 with 55.18\%. Among open-weights models, Kimi-K2.5 performs best with an overall accuracy of 51.63\%, followed by Qwen3-VL-32B-Thinking at 50.43\%, showing that competitive open-weights models can retain relatively strong performance under the more challenging EPD setting. Domain-wise strengths remain diverse in EPD: Gemini-3-Pro-Preview leads ML, PE, CM, and BH, Claude-Opus-4.5 performs best in EE, and GPT-5.2 (xhigh) achieves the top score in DA.

\subsection{Analysis of Reasoning Consistency}
As shown in Table~\ref{tab:arc}, to examine the relationship between error diagnosis and question‑answering ability, we construct two evaluation subsets based on model performance in the ETC task. For each model, we randomly select 200 samples where it correctly identified the error type and 200 samples where it misidentified the error type.
We then re-evaluate the same models on the original VQA task using only these two subsets. The results reveal a strong diagnosis–accuracy consistency: when the model previously diagnosed the error correctly, it also achieves significantly higher accuracy in answering the original visual question on the same subset. Conversely, samples that were misdiagnosed are strongly correlated with lower VQA accuracy.
This pattern indicates that a model’s ability to pinpoint what went wrong is closely tied to its underlying comprehension of the problem, which in turn supports more reliable answer generation in the original task.

\begin{table}[ht]
\centering
% \small
\begin{tabular}{lcc}
\toprule
\textbf{Model} & \textbf{Cor.} & \textbf{Incor.} \\
\midrule
Gemini-3-Pro-Preview        & 85.5 & 74.5 \\
GPT-5.2 (xhigh)               & 87.0 & 71.5 \\
Doubao-Seed-2.0-pro               & 85.0 & 72.0 \\
Qwen3-VL-32B-Instruct & 80.5 & 71.0 \\
LLaMA-4-Maverick      & 75.0 & 72.5 \\
\bottomrule
\end{tabular}
\caption{Experiments on original VQA accuracy (\%). For each model, ``Cor.'' is evaluated on a randomly sampled subset of 200 examples for which the model correctly identified the error type in ETC, while ``Incor.'' is evaluated on a randomly sampled subset of 200 examples for which the model incorrectly identified the error type.}
\label{tab:arc}
\end{table}

\subsection{Analysis of Multi-modal Alignment}
A key challenge in multi-modal reasoning is ensuring robust cross-modal alignment between visual inputs and textual descriptions~\cite{tang2025seam}. Inspired by \cite{DBLP:conf/iclr/NeoO0G0B25}, we selected samples from the ``Visual Perception Error'' category to  investigate why models succeed or fail in such cases. 
For the Qwen3-VL-32B-Instruct model, we perform a visual analysis by extracting the logit lens of each token at each layer after the text and image inputs are processed by the VLM.

\begin{figure}[!t]
    \centering
    \includegraphics[width=0.85\linewidth]{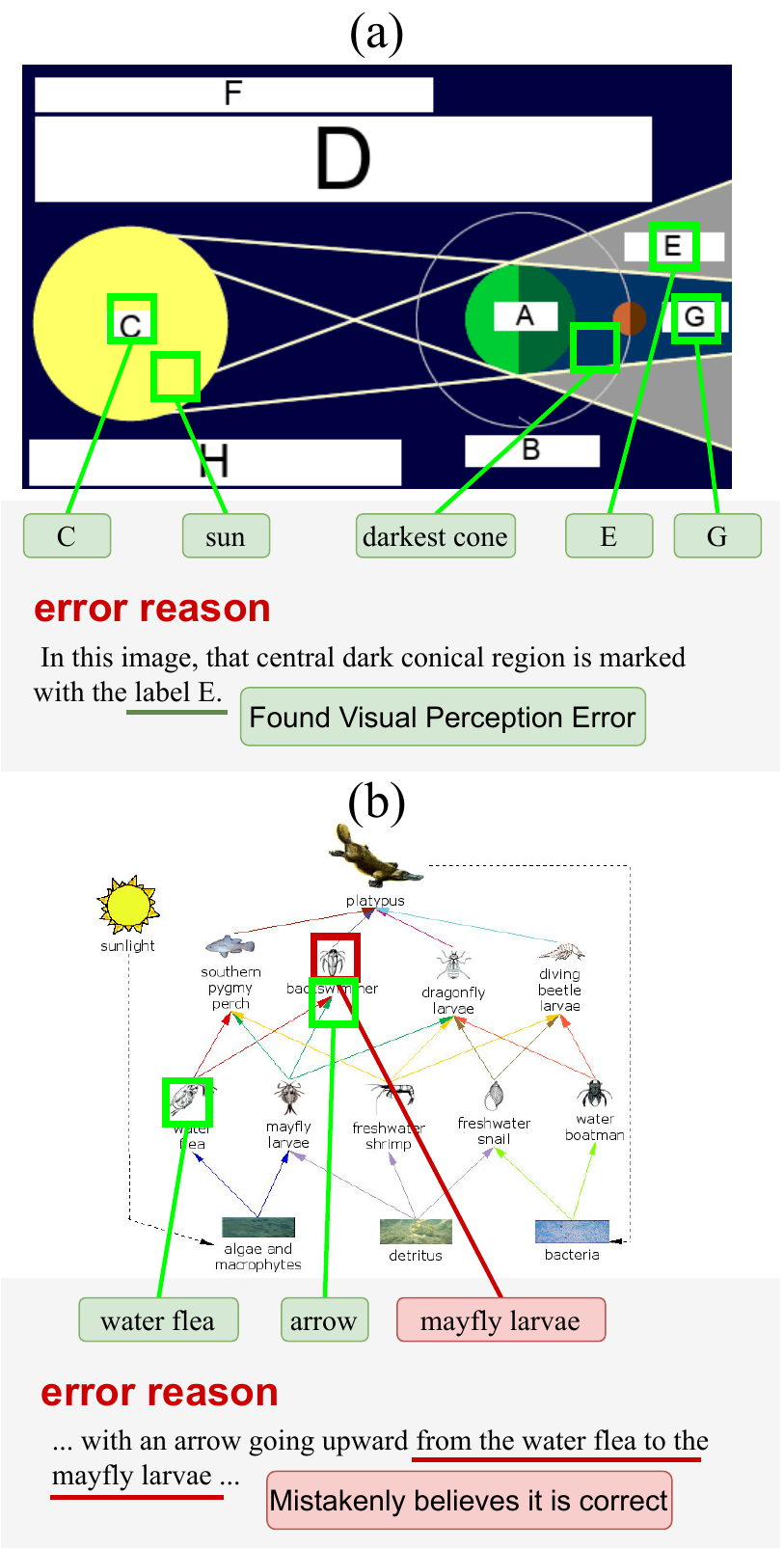}
    \caption{Logit lens visualization for image tokens. (a) Correct grounding. (b) Incorrect grounding.}
    \label{fig:activation_analysis}
    
\end{figure}

As shown in Figure \ref{fig:activation_analysis}, in case (a), where the model successfully identifies the error type, the relevant text tokens maintain a strong and correct semantic alignment with the corresponding image regions (e.g., the token ``darkest cone'' precisely attends to the visual cone area). 
In contrast, in case (b) where the model fails to detect the error, this alignment is disrupted. The model extracts irrelevant or ambiguous semantic information from the corresponding image patches (e.g., failing to associate the ``arrow'' token with its correct directional meaning relative to the objects).

\begin{table*}[t]
\centering
% \small
\setlength{\tabcolsep}{4pt}
\begin{tabular}{lcccc}
\toprule
\textbf{Model} & \textbf{VQA} & \textbf{VQA+Err} & \textbf{VQA+Err+StepKnow} & \textbf{VQA+Err+TypeKnown} \\
\midrule
Gemini-3-Pro-Preview        & 81.0 & 82.5  & 84.0 & \textbf{90.5} \\
GPT-5.2 (xhigh)               & 80.0 & 80.5  & 82.0 & \textbf{89.5} \\ 
Doubao-Seed-2.0-pro               & 80.5 & 81.5  & 83.0 & \textbf{88.5} \\
Qwen3-VL-32B-Instruct & 78.5 & 80.0  & 82.5 & \textbf{84.5} \\
LLaMA-4-Maverick      & 73.0 & 74.0  & 75.5 & \textbf{76.5} \\
% Qwen-VL-Max           & 75.5 & 76.0  & 77.5 & \textbf{80.5} \\
\bottomrule
\end{tabular}
\caption{Impact of error awareness on correction accuracy, evaluated on a randomly sampled subset of 200 examples from \model. \textbf{VQA} stands for the original VQA task, \textbf{Err} indicates that the model is additionally provided with an erroneous reasoning chain in the prompt, \textbf{StepKnown} specifies which step contains the error, and \textbf{TypeKnown} provides the exact error type classification.}
\label{tab:correction_analysis}
\end{table*}

\subsection{Exploration of Steps in Reasoning}
Prior research on error localization has predominantly focused on identifying which step in a reasoning chain contains an error. In this subsection, we go beyond step localization and examine how different levels of error awareness influence a model's ability to generate correct answers. 
We conduct this auxiliary analysis across multiple model families using a randomly sampled set of 200 examples from \model. These results are intended to compare different levels of error awareness under a controlled subset setting, rather than to be directly compared with the full-benchmark results in Table~\ref{tab:main_results}. 
As shown in Table~\ref{tab:correction_analysis}, we observe that merely exposing the model to the erroneous reasoning chain (\textit{VQA+Err}) yields almost no improvement over the baseline (\textit{VQA}). Annotating the erroneous step (\textit{VQA+Err+StepKnown}) results in a modest but consistent performance gain across all models. The most substantial improvement, however, occurs when the error type is provided (\textit{VQA+Err+TypeKnown}), leading to a clear and objective increase in correction accuracy.
Furthermore, we observe that the effectiveness of error-type guidance varies with model capability. For the strongest models in Table~\ref{tab:correction_analysis}, providing the precise error type yields the largest gains over both \textit{VQA+Err} and \textit{VQA+Err+StepKnown}, indicating that accurate diagnosis provides directly actionable information for answer correction.

% (a) \textbf{Success Case:} The model correctly identifies the error type. The visualization shows that the token representing the flawed concept (e.g., ``darkest cone'') is accurately grounded to the corresponding semantic region in the image. (b) \textbf{Failure Case:} The model fails to identify the error. As shown, the critical token maps to irrelevant visual semantics or fails to capture the specific spatial relationship, leading the model to incorrectly accept the flawed reasoning.

\section{Related Work}

\subsection{Evaluation of Multi-Modal Reasoning}
The rapid evolution of Vision-Language Models (VLMs) has necessitated rigorous benchmarks to measure their progress. Initial evaluations primarily focused on simple visual question answering (VQA). More recently, comprehensive benchmarks such as MMMU~\cite{yue2024mmmu}, MathVista~\cite{lu2023mathvista}, and MathVerse~\cite{zhang2024mathverse} have been introduced to evaluate complex reasoning capabilities across diverse domains, while related multimodal tasks have also expanded to temporally grounded video understanding~\cite{xu2025visulogic,ma2025ms}. However, these benchmarks typically adopt an outcome-oriented evaluation paradigm, focusing primarily on the correctness of the final answer. While high accuracy on these tasks suggests strong performance, it often creates an ambiguity: it is unclear whether the model genuinely understands the cross-modal content or is merely relying on superficial pattern matching. MMErroR departs from this tradition by shifting the focus from answer correctness to process-level verification. Instead of merely checking if the result is right, we evaluate whether the model can discern the validity of the reasoning path itself, providing a more transparent assessment of true multi-modal understanding.

\subsection{Hallucination and Visual Consistency}
Ensuring the reliability of VLMs has led to a significant body of work on hallucination detection, mitigation, and mechanism analysis. Benchmarks like POPE~\cite{li2023evaluating} and HallusionBench~\cite{guan2024hallusionbench}, together with related studies on fine-grained visual perception, causal attention, shallow-layer attention repair, and attention-sink analysis~\cite{ma2025fine,zhao2025mca,zhang2025shallow,zhang2026drives}, have been instrumental in understanding and reducing object-level failures, such as errors in object existence or attribute description. While these works effectively target Visual Perception Error, they often overlook the complexity of higher-order cognitive failures. Multi-modal reasoning requires not only accurate perception but also the logical integration of visual data with parametric knowledge. As defined in our taxonomy, errors can stem from diverse sources beyond perception, including Visual Perception Error (VPE), Knowledge Deployment Error (KDE), Reasoning Error (RE), and Question Comprehension Error (QCE). MMErroR provides a broader coverage of these failure modes, requiring models to identify errors in logic and factual application, not just in visual grounding.

\subsection{Error Localization and Erroneous Reasoning}
Recent research has begun to scrutinize the intermediate steps of reasoning and the relevance of intermediate evidence to better diagnose model failures~\cite{ruan2025vlrmbench,imani2026trace}. Existing benchmarks~\cite{fang2025flux,yan2024errorradar} represent a shift towards evaluating step-by-step consistency. These existing benchmarks primarily focus on Error Localization, identifying which step in a sequence is incorrect. While localization is useful, it offers limited insight into why the model failed. MMErroR distinguishes itself by enforcing ETC. We argue that a robust VLM must be capable of introspective diagnosis: determining whether a failure was caused by misinterpreting a diagram, applying the wrong formula, or a logical fallacy. Furthermore, beyond ETC, MMErroR also includes an EPD task, which requires models to explicitly invoke an error label before attribution under controlled erroneous conditions.

\section{Conclusion}
In this paper, we introduce \model, a novel fine-grained benchmark designed to evaluate the reasoning capabilities of VLMs by shifting the evaluation paradigm from final-answer correctness to process-level error diagnosis.
\model contains \num samples spanning 24 subdomains across six top-level domains and supports two core evaluation tasks: Error-Type Classification and Error Presence Detection.
The current release is a controlled diagnostic stress-test built on reasoning chains with exactly one verified root-cause error.
Through evaluation of 12 representative VLMs, we find that even the strongest models exhibit significant limitations in identifying and classifying reasoning errors, with the top performer achieving only 66.65\% overall accuracy.
We hope \model can stimulate further research toward building more reliable and interpretable multi-modal reasoning systems, while verified clean chains and multi-error cascades remain important future extensions.

% \clearpage
\section*{Limitations}
Despite the comprehensive design of MMErroR, several limitations remain. First, our benchmark is constructed such that each sample contains a single, coherent reasoning error. While this isolation allows for precise diagnostic attribution, real-world reasoning failures often involve cascading or multiple simultaneous errors, which are not currently modeled in this dataset. Second, while we employ a rigorous multi-stage human verification process to ensure correctness and quality, the initial erroneous reasoning chains are generated via model-assisted synthesis. This reliance may introduce subtle biases in error patterns or linguistic styles specific to the generator model. 
Future work may explore open-ended generation metrics and multi-error scenarios to address these gaps.

\section*{Acknowledgments}
This research was supported in part by National Science and Technology Major Project (2021ZD0111502), Natural Science Foundation of China (U24A20233, 62406078), CCF-DiDi GAIA Collaborative Research Funds (CCF-DiDi GAIA 202521), and Guangdong Laboratory of Artificial Intelligence and Digital Economy (SZ)(GML-KF-24-23).


\bibliography{custom}

\appendix
\clearpage

\section{Complexity-Aware Filtering}
\label{sec:appendix_1}
We quantify question difficulty with a lightweight feature vector:
\begin{itemize}
  \item comparative tokens (\texttt{<}, \texttt{>}, \texttt{taller}, \texttt{heavier})
  \item negations (\texttt{not}, \texttt{never}, \texttt{except})
  \item numerical quantities
  \item open-ended wh-words (\texttt{why}, \texttt{how many steps})
  \item presence of domain-specific formulas (regex match)
\end{itemize}
Each feature is z-scored and equally weighted into a single complexity score:
\[
\text{score} = \frac{1}{5}\sum_{i=1}^{5} z_{i}.
\]

To over-sample harder instances while preserving medium-easy diversity, we fit a Gaussian $\mathcal{N}(\mu, \sigma^{2})$ over all scores and draw 10\,000 samples from the upper-half tail $(\mu + 0.5\sigma,\; \mu + 2\sigma)$. This raises the mean complexity from $0.00$ to $+0.82$ while retaining few lower-complexity items for evaluation robustness.

\section{Prompt Template}
\label{sec:appendix_2}
To ensure transparency and reproducibility in constructing \model, we detail here the prompts used to generate erroneous reasoning chains. Each prompt is carefully designed to elicit plausible yet incorrect reasoning while maintaining linguistic coherence and contextual relevance.

\begin{lstlisting}[style=promptstyle]
======================
ERROR TAXONOMY (CHOOSE EXACTLY ONE)
======================
1. A_Visual_Perception_Error
   The model makes a mistake in visually interpreting the image, such as:
   - Misreading text or numbers (OCR error, e.g., reading "1.0" as "10").
   - Misinterpreting chart or table values (e.g., confusing bar heights).
   - Confusing colors, shapes, positions, or object counts.
   - Mislocating objects (e.g., assigning the wrong label to a region).
   The *reasoning logic itself* (once the wrong visual input is assumed) should be mostly correct.

2. B_Reasoning_Error
   The model correctly perceives the visual information but makes a mistake in:
   - Arithmetic or calculation (e.g., 3 + 4 + 5 = 11).
   - Combining quantities, units, or proportions.
   - Logical deduction or step-by-step reasoning.
   All visually extracted facts should be correct; the error is in the mental steps.

3. C_Question_Comprehension_Error
   The model understands the image reasonably well but misinterprets the question, such as:
   - Answering a different but related question.
   - Ignoring constraints (e.g., "only red objects", "in the last row").
   - Mixing up entities asked about (e.g., answering about Bob when asked about Alice).
   - Answering about a subset or superset instead of the exact target.
   The reasoning may be logically consistent, but it is applied to the wrong interpretation of the QUESTION.

4. D_Knowledge_Deployment_Error
   The model sees the image correctly and understands the question, but:
   - Uses the wrong external knowledge (e.g., incorrect physical or scientific fact).
   - Misapplies a known formula or concept.
   - Retrieves or applies an irrelevant or incorrect fact not supported by the image.
   Visual perception and question understanding should be correct; the error comes from using the wrong background knowledge or formula.

======================
TASK
======================
Given IMAGE, QUESTION, and CORRECT_ANSWER:

1. Carefully inspect the IMAGE and QUESTION.
2. Decide which single error type (A, B, C, or D) can produce a **realistic and plausible** incorrect answer.
3. Construct a natural, confident reasoning chain that:
   - Uses the visual information.
   - Leads to an incorrect final answer.
   - Contains **exactly one** of the error types above.
   - Is otherwise as correct and detailed as possible.
4. Do **NOT** explicitly say that you are making an error, simulating a failure, or referring to labels or taxonomy.
   - Write as if you are a normal LVLM answering the QUESTION.
5. Ensure the final predicted answer in `error_reason` is **different from** CORRECT_ANSWER.
6. Set `label` to exactly one of:
   - "A_Visual_Perception_Error"
   - "B_Reasoning_Error"
   - "C_Question_Comprehension_Error"
   - "D_Knowledge_Deployment_Error"
\end{lstlisting}

\begin{table}[t]
\centering
% \small
\resizebox{1\linewidth}{!}{
\begin{tabular}{lcccc}
\toprule
\textbf{Model} & \textbf{0-shot} & \textbf{1-shot} & \textbf{2-shot} & \textbf{4-shot} \\
\midrule
Gemini-3-Pro-Preview        & 66.5 & 67.0 & 67.5 & 68.5 \\
Doubao-Seed-2.0-pro               & 65.0 & 66.5 & 66.5 & 67.0 \\
Qwen3-VL-32B-Instruct & 49.5 & 53.0 & 55.0 & 56.0 \\
LLaMA-4-Maverick      & 39.5 & 43.5 & 45.5 & 47.0 \\
\bottomrule
\end{tabular}
}
\caption{Impact of ICL on the ETC task, evaluated on a randomly sampled subset of 200 examples from \model. All 0-shot and few-shot results in this table are obtained under the same sampled subset and prompt template, and are therefore not directly comparable to the full-benchmark results in Table~\ref{tab:main_results}.}
\label{tab:icl_exploration}
\end{table}

\section{Few-shot Learning Exploration}
We explore whether self-oversight capabilities can be elicited or improved via In-Context Learning (ICL)~\cite{brown2020language} and few-shot prompting. To this end, we conduct an auxiliary experiment on a randomly sampled subset of 200 examples from \model, and test 0-shot, 1-shot, 2-shot, and 4-shot prompts across various models, as shown in Table~\ref{tab:icl_exploration}.

% \section{Additional ETC Confusion Matrices}

% Figure~\ref{fig:etc-cm-balanced12} shows the row-normalized confusion matrices of 12 representative models on the ETC task. We restrict the visualization to the four error types (KDE, VPE, RE, and QCE), and normalize each row to sum to 1 to remove the direct effect of class frequency. Overall, the non-KDE rows are not dominated by the KDE column, and the diagonal entries for the minority classes (especially RE and QCE) remain relatively high. This indicates that the models do not exhibit strong majority-class collapse, and that our main conclusions are not simply artifacts of label imbalance.

% \begin{figure*}[t]
%   \centering
%   \includegraphics[width=\textwidth]{figure/etc_cm_balanced12_doublecol_4x3.pdf}
%   \caption{Row-normalized confusion matrices for 12 representative models on the ETC task. Each panel shows a 4$\times$4 confusion matrix over the four error types (KDE, VPE, RE, QCE), with rows corresponding to gold labels and columns to model predictions. Rows are normalized to sum to 1, so each cell gives the conditional distribution of predictions given the true error type. The non-KDE rows are not dominated by the KDE column, and the diagonal entries for the minority classes (RE and QCE) remain relatively high, indicating that models do not exhibit strong majority-class collapse and that our conclusions are not driven solely by label imbalance.}
%   \label{fig:etc-cm-balanced12}
% \end{figure*}

\end{document}